\documentclass{esannV2}
\usepackage[dvips]{graphicx}
\usepackage[latin1]{inputenc}
\usepackage{amssymb,amsmath,array}
\usepackage{subfigure}

\def \rb {{\rangle}}
\def \lb {{\langle}}
\def \emptyz {{0\!\!\!/}}

\begin{document}
\title{Geometric Models with Co-occurrence Groups}


\author{Joan Bruna$^{1,2}$ and St\'ephane Mallat$^2$
%
%
\vspace{.3cm}\\
%
1- Zoran France \\
8/16 rue Paul Vaillant Couturier, 92240, Malakoff - France
\vspace{.1cm}\\
2- Ecole Polytechnique - CMAP \\
Route de Saclay, 91128 Palaiseau - France
%
}

\maketitle

\begin{abstract}
A geometric model of sparse signal representations is introduced
for classes of signals. It is computed by optimizing co-occurrence
groups with a maximum likelihood estimate calculated with a Bernoulli
mixture model. Applications to face image compression and 
MNIST digit classification
illustrate the applicability of this model.
\end{abstract}

\section{Introduction}

Finding image representations with a dimensionality reduction 
while maintaining relevant information for classification, 
remains a major issue.
Effective approaches have recently been developed based
on locally orderless representations as proposed by Koendering and Van Doom
\cite{Koenderink}. They observed that 
high frequency structures are important for recognition but
do not need to be precisely located. This idea has inspired a
family of descriptors such as SIFT \cite{lowe} or HOG \cite{Trigg},
which delocalize the image information over large neighborhoods,
by only recording histogram information. 
These histograms are usually computed over wavelet like 
coefficients, providing a multiscale image representation with 
several wavelets having different orientation tunings. 
%

This paper introduces a new geometric image representation 
obtained by grouping coefficients that have co-occurrence properties across
an image class. It provides a locally 
orderless representation where sparse descriptors are 
delocalized over groups 
which optimize the coefficient co-occurrences, and can 
be interpreted as a form of parcellization \cite{flandin}.
Section \ref{geom-sec} reviews wavelet image representations and
the notion of sparse geometry through significant sets.
Section \ref{mixt-sec} introduces our co-occurrence grouping model
which is optimized with a maximum likelihood approach.
Groups are computed
from a training sequence in Section \ref{Bandfosec}, 
using a Bernoulli mixture approximation.
Applications to face image compression are
shown in Section \ref{comp-sec} and the application of this
representation is illustrated for MNIST image 
classifications in Section \ref{MNIST-sec}.

\section{Geometric Significance Set}
\label{geom-sec} 

Sparse signal representations are obtained by decomposing signals over
bases or frames $\{ \phi_p \}_{p \in \bar y}$ which take 
advantage of the signal regularity to produce
many zero coefficients. 
A sparse representation 
is obtained by keeping 
the significant coefficients above a threshold $T$, 
$$y = \{p \in \bar y~~:~~|\lb f , \phi_p \rb| > T~ \}$$
The original signal can be 
reconstructed with a dual family 
$f = \sum_{p \in \bar y} \lb f , \phi_p \rb \, \tilde \phi_p$, 
and the resulting sparse approximation is
$f_y = \sum_{p \in y} \lb f , \phi_p \rb \, \tilde \phi_p$.

Wavelet transforms
compute signal inner products with 
several mother wavelets $\psi^d $
having a specific direction tuning, and which are
dilated by $2^j$ and translated by $2^j n$:
$\phi_p = \psi^d_{j,n}$.
Separable wavelet bases are obtained with $3$ mother wavelets \cite{Mallatbook}, 
in which case the total number $|\bar y|$ 
of wavelets is equal to the image size. 

Let $|y|$ be the cardinal of the set $y$.
In absence of prior information on $y$, the number of bits
needed to code $y$ in $\bar y$ is 
$R_0 = \log_2 \binom {|\bar y|} {|y|}$ .

One can also verify \cite{Mallatbook}
that the number of bits required to encode
the values of coefficients in $y$ is proportional to $|y|$
and is smaller than $R_0$ so that the coding budget is indeed dominated by
$R_0$ which carries most of the image information. 

\section{Co-occurrence Groups}
\label{mixt-sec} 

In a supervised classification problem, a geometric model
defines a prior model of
the probability distribution $q(y)$. 
There is a huge number $2^{|\bar y|}$ of 
subsets $y$ in $\bar y$. Estimating the probability
$q (y)$ from a limited training set thus requires using a
simplified prior model.

A signal class is represented by a random vector  whose realizations
are within the class and whose significance sets $y$ are
included in $\bar y$. A mixture model is introduced
with co-occurrence groups $\theta(k)$ of constant size $s$,
which define a partition of the overall index set $\bar y$
\[
\bar y = \cup_{k} \theta(k)~~\mbox{with}~~ |\theta(k)| = s~~
\mbox{and}~~ \theta(k) \cap \theta(k') = \emptyz 
~~\mbox{if}~~ k \neq k'~.
\]

Co-occurrence groups $\theta(k)$ 
are optimized by enforcing that all coefficients
have a similar behavior in a group and hence that 
$y \cap \theta(k)$ is either almost empty or almost equal to 
$\theta(k)$ with a high probability.
The mixture model assumes that the distributions of the components
$y \cap \theta(k)$ are independent.
The distribution $q(y \cap \theta(k))$
is assumed to be uniform 
among all subsets of $\theta(k)$ of cardinal
$z(k) = |y \cap \theta(k)|$. Let $q_k (z(k))$ be its distribution,
\[
q (y|\theta) = \prod_k q(y \cap \theta(k)) = 
\prod_k q_k (z(k)) \binom{s} {z(k)}^{-1}
\]
This co-occurrence model is identified with a maximum log-likelihood
approach which computes
\[
\arg \max_{\theta}  \sum_k 
\Bigl( - \log_2 {\binom {s} {z(k)}} + \log_2 q_k (z(k)) \Bigr)~.
\]
\section{Group Estimation with Bernoulli Mixtures}
\label{Bandfosec}

Given a training sequence of images $\{ f_l \}_{l \leq L}$ that belong to
a class, we optimize the group co-occurrence by approximating the
maximum likelihood with a Bernoulli mixture.

Let $y_l$ be the significant set of $f_l$.
The log likelihood is calculated with
\begin{equation}
\label{conasdnwosn}
{\cal L} (y,\theta) = \sum_l \sum_k 
\Bigl( - \log_2 {\binom {s} {z_l(k)}} + \log_2 q_k (z_l(k)) \Bigr)~~
\mbox{with}~~z_l (k) = |y_l \cap \theta(k)|~.
\end{equation}
The maximization of this expression is obtained using the
 Stirling formula which approximates the first term by the
entropy of a Bernoulli distribution.
Let us write
$q_{k,l} (0) = z_l (k)/s$ and $q_{k,l} (1) = 1- z_l (k)/s$,
the Bernoulli probability distribution associated to $z_l (k)/s$. 
Let us specify 
the groups $\theta(k)$ by the inverse variables $k(p)$ 
such that $p \in \theta(k(p))$. It results that
\begin{eqnarray*}
\sum_k - \log_2 {\binom   {s} {z_l(k)}} &\approx &
z_l (k) \, \log_2 \left(\frac{z_l(k)}{s}\right) + (s - z_l (k)) \log_2 \left(1 - \frac{z_l (k)}{s}\right) ~\\
&=& \sum_{p \in \bar y} \log_2 q_{k(p),l} (1_{y_l} (p))~.
\end{eqnarray*}

The distribution $q_k$ is generally unknown and must therefore be estimated.
The estimation is regularized by approximating this distribution with a 
piecewise constant distribution $\hat q_k$
over a fixed number of quantization bins,
that is small relatively to the number of realizations $L$. 
The likelihood (\ref{conasdnwosn}) is thus
approximated by a likelihood over the Bernoulli mixture, which
is optimized over all parameters:
\begin{equation}
\label{conasdnwosn3}
\arg \min_{\theta, z_l, \hat q_k} 
- \sum_l (\sum_{p \in \bar y} \log q_{k(p),l} (1_{y_l} (p)) + 
\sum_k \log_2 \hat q_k (z_l (k)) )~.
\end{equation}
The following algorithm, 
minimizes (\ref{conasdnwosn3}) by updating separately
the Bernoulli parameters
$z_l (k)$, the distribution $\hat q_k$ and the grouping variables
$k(p)$. 

The minimization algorithm begins with a random initialization of 
groups $\theta(k)$ of same size $s$. 
The empirical histograms $\hat q_k$ are initialized to uniform distributions. The algorithm iterates the
following steps:
\begin{itemize}
\item Step 1: Given $\{ \theta(k) \}_k$ and $\{ \hat q_k \}_k$
compute $\{ z_l (k) \}_{l,k}$ which minimizes (\ref{conasdnwosn3}) by
minimizing
\begin{equation}
\label{asnfwonds}
- \log_2 \hat q_k (z_l(k)) -
z_l (k) \, \log_2 \frac{z_l(k)} s - (s - z_l (k)) \log_2 (1 - \frac{z_l (k)} s)) ~.
\end{equation}

\item Step 2: Update $\{ \hat q_k \}_k$ to minimize (\ref{conasdnwosn3}) 
as the normalized histogram of the 
updated parameters $\{ z_l (k) \}_l$ over a predefined number of bins. 

\item Step 3: Update the group indexes  $\{ k(p) \}_{p}$ to minimize
(\ref{conasdnwosn3}) by minimizing
\begin{equation}
\label{asnfwonds2}
 -\sum_{p \in \bar y } \log q_{k(p),l} (1_{y_l} (p))~,
\end{equation}
for groups of constant size $|\theta(k)| = s$. 
%
\end{itemize}
This algorithm is guaranteed to converge to a local maxima
because each step further increases the log-likelihood. In fact, it is the equivalent
of the $K$-means algorithm adapted to the mixture model considered here.

\section{Face Compression}
\label{comp-sec}

\begin{figure}[h!]
\centering
\subfigure[] {\includegraphics[scale=0.25]{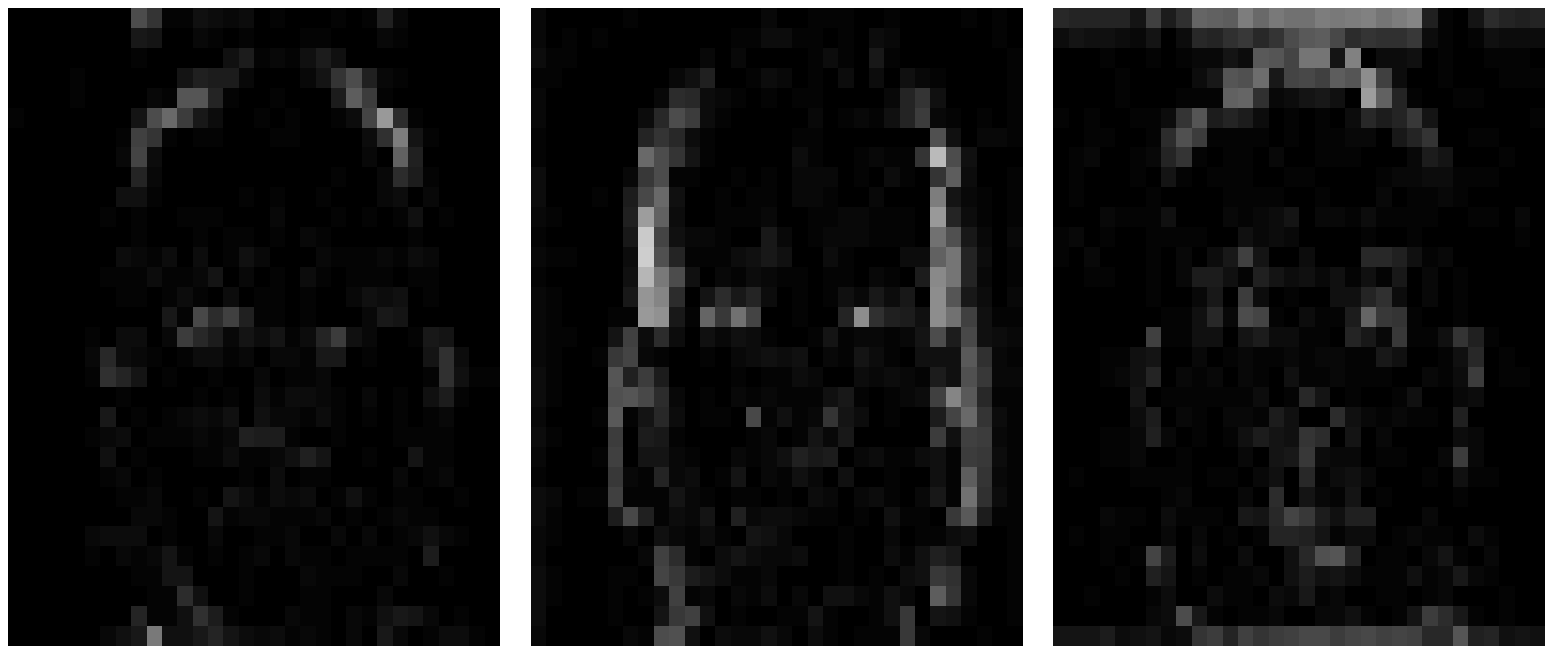}}
\subfigure[] {\includegraphics[scale=0.25]{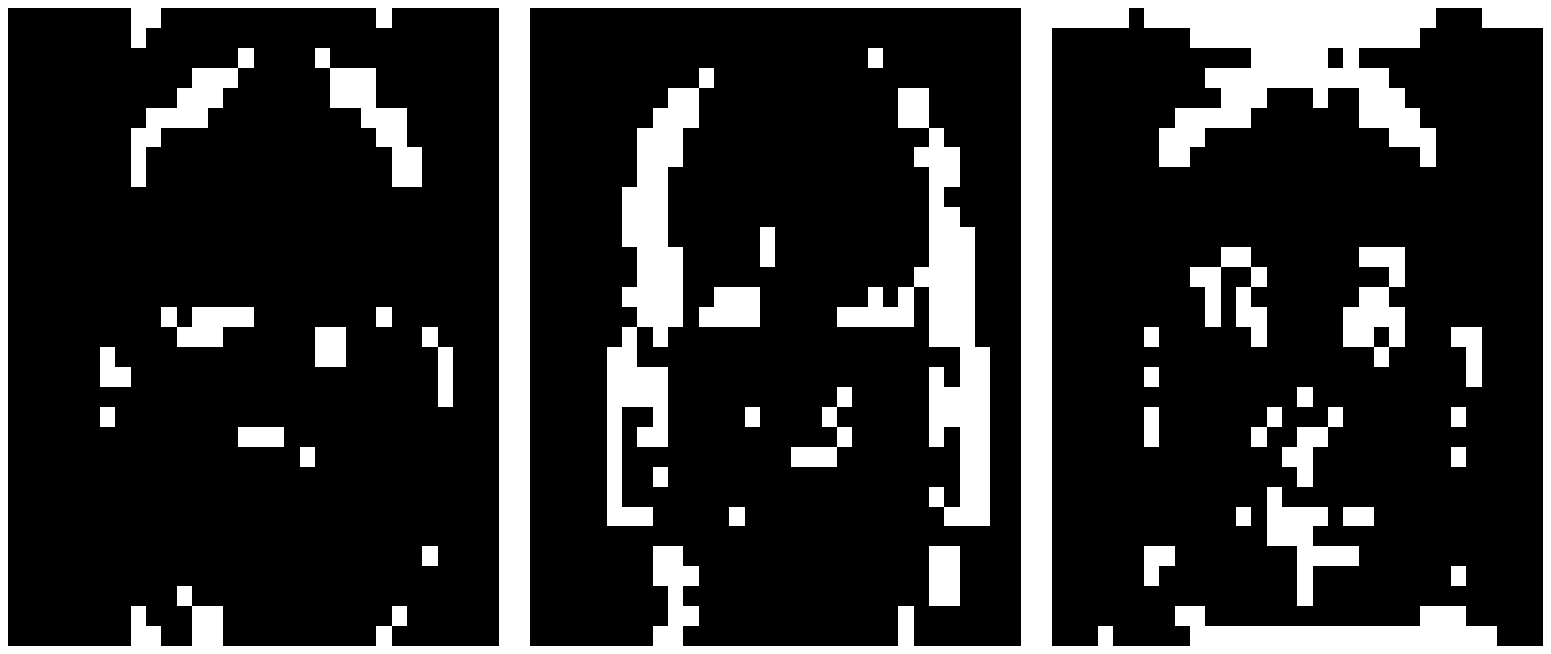}}
\subfigure[] {\includegraphics[scale=0.25]{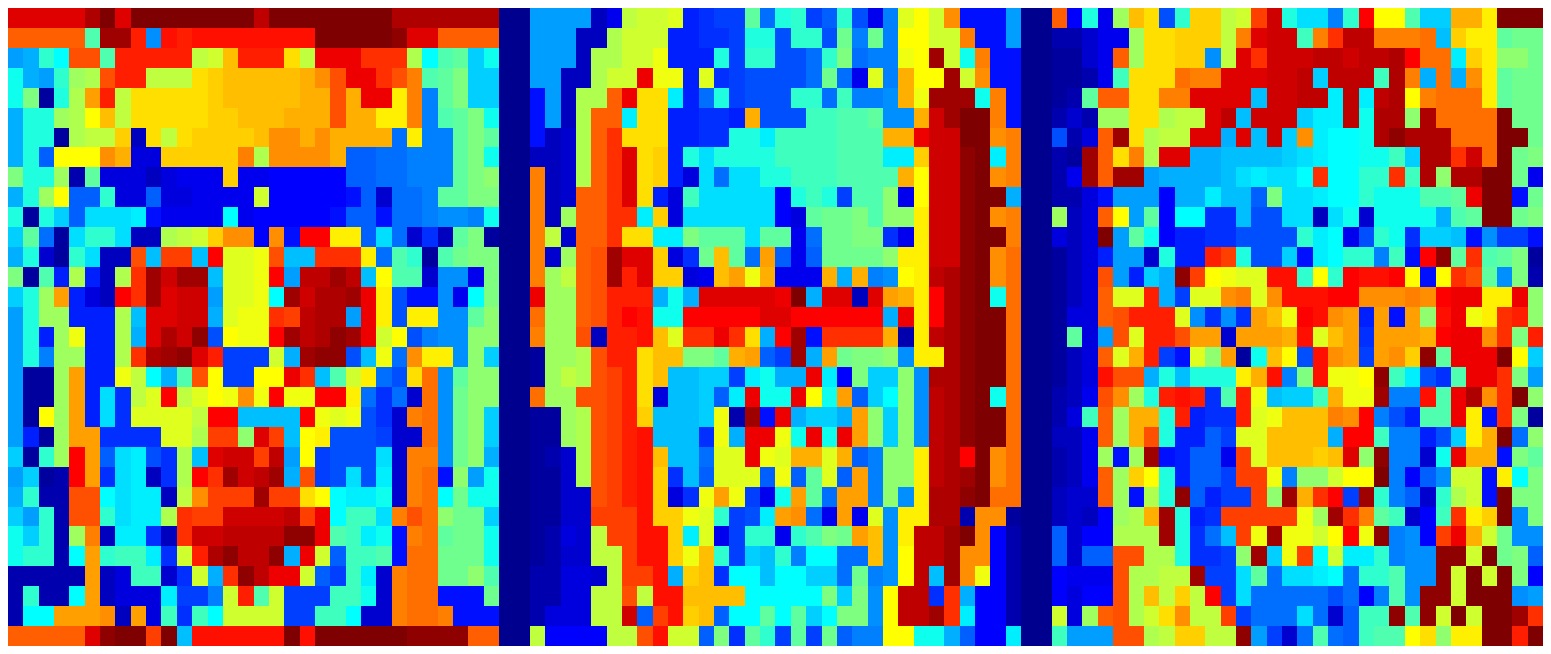}}

\caption{(a): Images of wavelet coefficients $|\langle f , \psi^d_{j,n} \rangle|$ for three
directions $d=1,2,3$ at a scale $2^j = 2^2$
(b): thresholded coefficients, defining the significance maps $y_l$. (c): grouping obtained with optimal 
group size $s=16$. The stable geometric features are clearly visible.}\label{cooc}

\end{figure}

\begin{figure}[h!]
\centering
\includegraphics[scale=0.25]{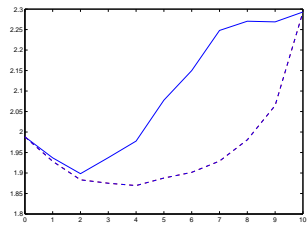}
\caption{\textit{Solid}: bit rate using fixed 
square groups of size $s$ as a function of  $\log_2 s$. 
\textit{Dashed}: bit rate (equal to minus the log likelihood, in bits per pixel) 
using the optimal groups of size $s$ as a function of $\log_2 s$.}\label{bitrate} 
\end{figure}

To illustrate the efficiency of this grouping strategy, it is first applied
to the compression of face images that have been approximately registered.
A database of 170 face images were used for training and a different
set of 30 face images were used for testing.
Figure \ref{cooc} shows the optimal
co-occurrence groups obtained over wavelet coefficients by applying
the maximum log-likelihood algorithm on the training set. 
The encoding cost of the significance map 
using the optimized model is equal to minus the log-likelihood
of this model. 
Figure \ref{bitrate} shows the evolution of the average bit budget needed
to encode the significance maps with the Bernoulli mixture over 
optimized co-occurrence groups, depending upon the groups size $s$. 
The optimal group size
which maximizes the log-likelihood and hence minimizes the encoding
cost over all group sizes is $s = 16$.  



When $s$ is equal to the image size, there is a single group and the 
encoding is thus equivalent to a standard image coding using no prior 
information on the class. The bit rate is also compared with a Bernoulli
mixture computed with a partition into square groups $\theta(k)$, 
as a function 
of $s$. Figure \ref{bitrate} shows that the optimized co-occurrence grouping
improves the bit rate  by 20 \% relatively to the case where there
is a single group, and also with respect to the fixed square groups, which means that the
optimal grouping 
provides a geometric information which is stable across the image class. The optimal group size 
$s = 16$ also 
gives an estimation of the image deformations that are due to 
variations of scaling and eye positions and to intrinsic variations of
faces in the database.

\section{Random MNIST Digit Classification}
\label{MNIST-sec}

This section shows the classification ability of our geometric 
representation despite the presence of strong variability in the images.
The test is performed using the standard MNIST database of digits. 
This database is relatively simple and without any modification
of the image representation an SVM classifier can reach $1.4\%$ of error
with a training set of 60,000 images. This section 
shows that our geometric co-occurence model can learn with much less
training elements and for more complex images.

\begin{figure}
     \centering
          \includegraphics[width=.22\textwidth]{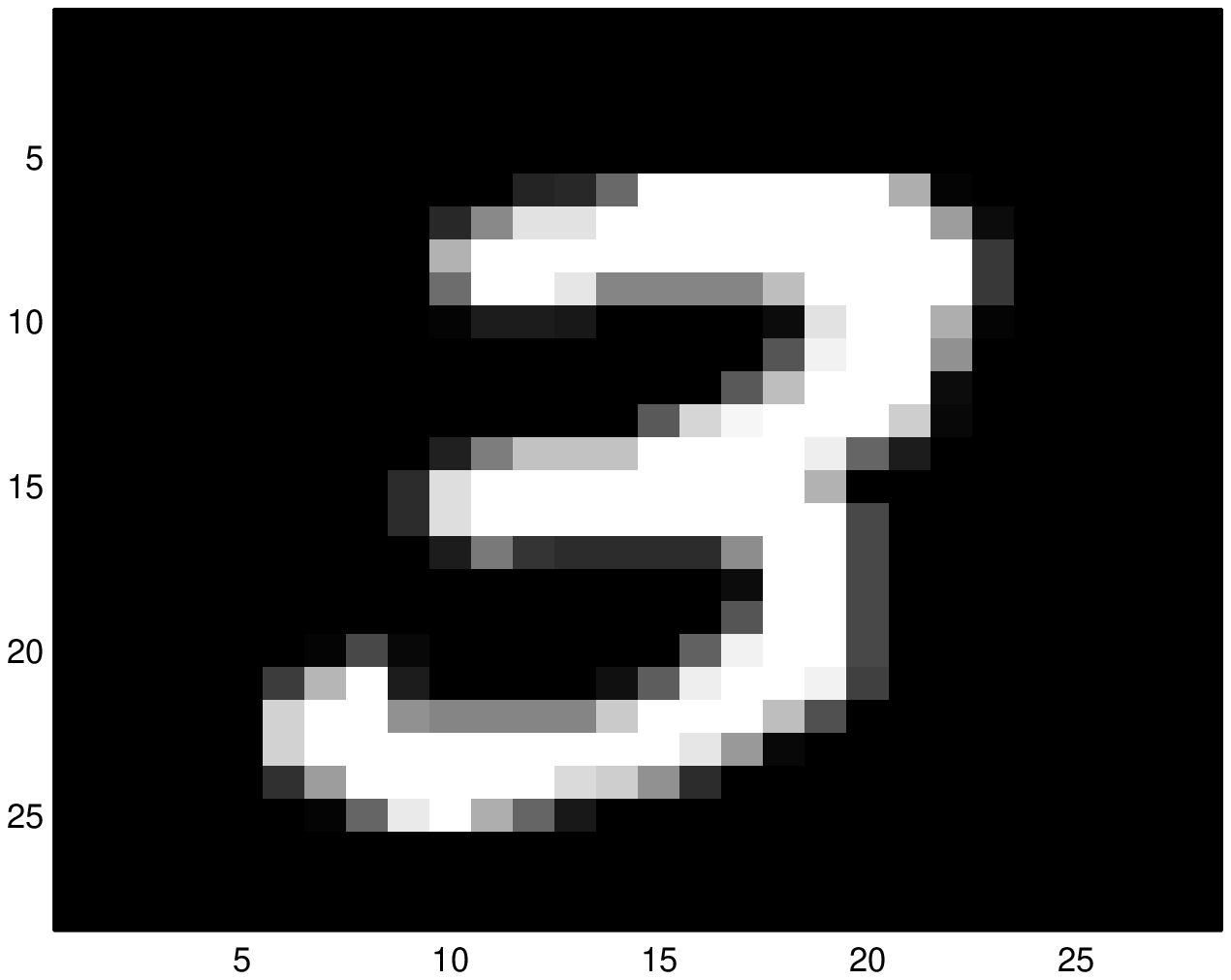}
          \includegraphics[width=.22\textwidth]{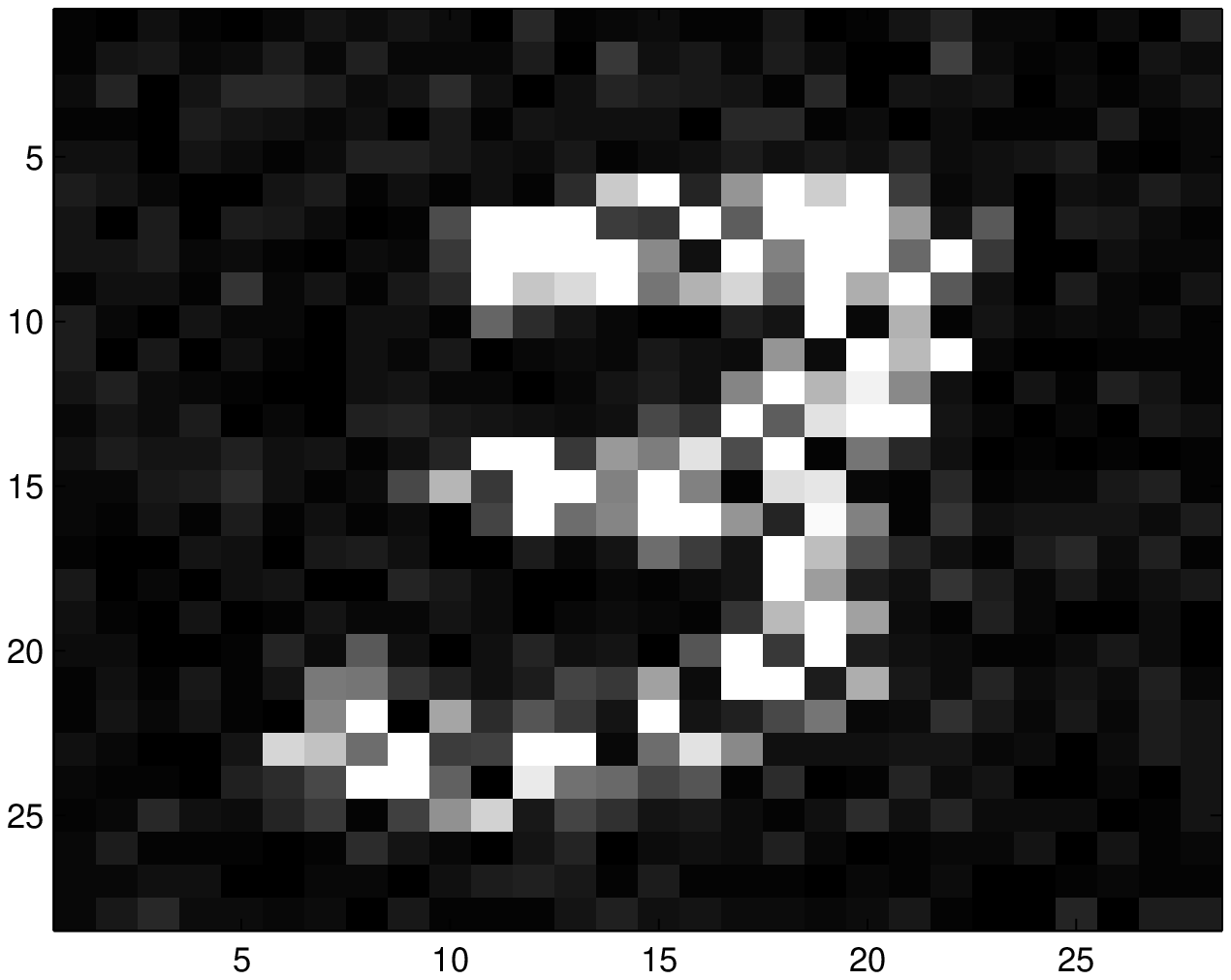}
           \includegraphics[width=.22\textwidth]{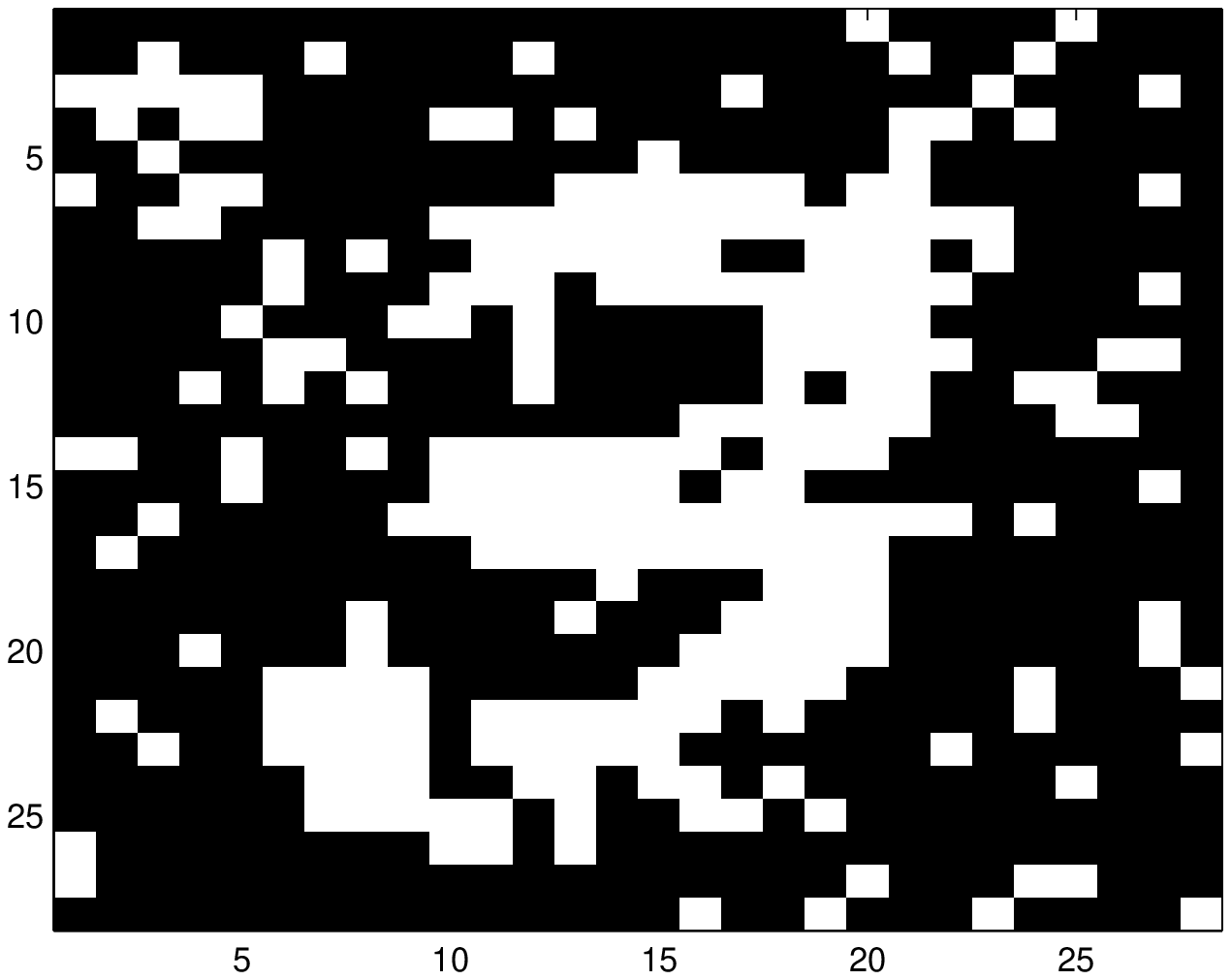}
           \includegraphics[width=.22\textwidth]{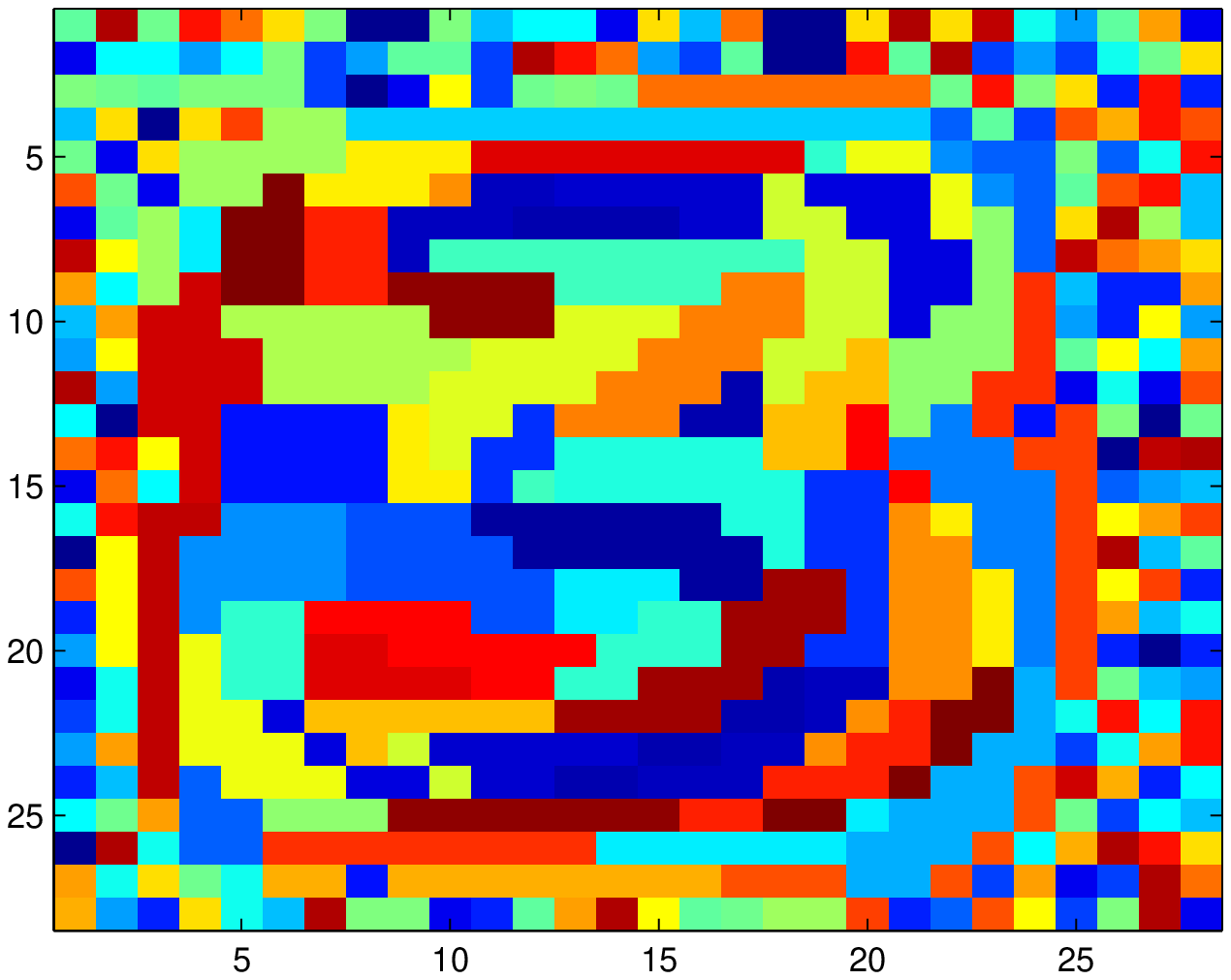}
     \caption{Digit example. \textit{From left to right}: original digit taken from MNIST database, random digit, 
     significance maps, and grouping obtained by using the described algorithm.}
     \label{mnist_figure}
\end{figure}

To take into account texture variation phenomena, which are a central
difficulty for geometric models, a white noise texture is introduced.
A digit image $f[n]$ is transformed into a random digit
$\tilde f[n] = (f[n] + C) W[n]$
where $W[n]$ is a normalized Gaussian white noise.
The significance maps of these digits are simply obtained with a
thresholding as shown in Figure \ref{mnist_figure}. It yields a binary image with a low
density binary texture on the digit background and high density texture
on the digit support. Visually, the digit is still perfectly recognizable
despite the texture variability. With $4000$ training images
an SVM with a polynomial kernel yields a very low recognition
rate of \textbf{21\%} on a different set of $10000$ test images. 

Figure \ref{mnist_figure} shows the optimal co-occurrence groups of size $14$ computed
with the minimization algorithm of Section \ref{Bandfosec}. Despite the geometric
variability, the algorithm is able to extract co-occurrence groups that
do correspond to the digit structures and their deformations.
To each digit $0 \leq d \leq 9$, corresponds an optimized 
co-occurrence grouping $\theta_d$. 
Let ${\cal L}(y,\theta_d)$ be the likelihood
of the significance map $y$ of $f$ with the grouping model $\theta_d$.
An SVM classifier is
trained on the feature vector $\{{\cal L} (y,\theta_d(k) ) \}_{0 \leq d \leq 9,0\leq k \leq K}$
, of dimension $10\cdot 56$ with groups of size $14$.
With $4000$ training images this classifier yields a recognition
rate of \textbf{9\%} on a different set of $10000$ test images. 
A simple maximum likelihood classifier (MAP) associates
to each test image $f$ the digit class
\[
\tilde d = \arg \max_{0 \leq d \leq 9} {\cal L} (f,\theta_d)~.
\]
With $4000$ training examples, 
this simple classifier yields a recognition rates of 18\% for random
digits, which is already better than the SVM applied on the original pixels.
\section{Conclusion}
This paper introduces a new approach to define the geometry of a class
of images computed over a sparse representation, using co-occurrence groups.
These co-occurrence groups are computed with a maximum log likelihood
estimation calculated over optimized Bernoulli mixture model. An
algorithm is introduced to optimize the group computation. The application
to face image compression shows the efficiency of this encoding approach,
and the ability to compute co-occurrence groups that provide 
stable information across the class. A classification test is performed
over textured digits, which shows that the algorithm can take into account
texture geometry and provide much better classification rates than a
standard pixel based image representation.

\bibliographystyle{plain}

\end{document}